\documentclass{article}
\usepackage{spconf,amsmath,graphicx}
\usepackage{amsfonts}
\usepackage{mathtools}
\usepackage{subfig}
\usepackage{verbatim}
\usepackage{xcolor}
\usepackage{url}
\title{Improving Semi-supervised End-to-end Automatic Speech Recognition using CycleGAN and Inter-domain Losses}
\name{Chia-Yu Li and Ngoc Thang Vu}
\address{
  Institute for Natural Language Processing (IMS), University of Stuttgart, Germany}
\copyrightnotice{978-1-6654-7189-3/22/\$31.00~\copyright2023 IEEE}
\begin{document}
%\ninept
%
\maketitle
\begin{abstract}
We propose a novel method that combines CycleGAN and inter-domain losses for semi-supervised end-to-end automatic speech recognition. Inter-domain loss targets the extraction of an intermediate shared representation of speech and text inputs using a shared network. CycleGAN uses cycle-consistent loss and the identity mapping loss to preserve relevant characteristics of the input feature after converting from one domain to another. As such, both approaches are suitable to train end-to-end models on unpaired speech-text inputs. In this paper, we exploit the advantages from both inter-domain loss and CycleGAN to achieve better shared representation of unpaired speech and text inputs and thus improve the speech-to-text mapping. Our experimental results on the WSJ eval92 and Voxforge (non English) show $8\sim8.5\%$ character error rate reduction over the baseline, and the results on LibriSpeech test\_clean also show noticeable improvement. 
\end{abstract}
\section{Introduction}
End-to-end (E2E) automatic speech recognition (ASR) directly learns the mapping from acoustic feature sequence to a label, character or subword, sequence using a encoder-decoder architecture \cite{e2e-rnn-asr-2014,e2e-rnn-asr-2014-firstresult,deepspeech-2014,e2e-rnn-wfst-asr-2015,e2e-attention-asr-2015,e2e-attention-asr-2016,las-2016,on-online-attention-e2e-2016,on-training-rnn-e2e-2016,very-deep-cnn-e2e-asr-2017,e2e-lstm-asr-2017,jointlyc2catt,hybride2e}. One of the popular architectures is hybrid CTC/attention, which effectively utilizes the advantages of connectionist temporal classification (CTC) based model and attention based model in training and decoding \cite{jointlyc2catt,hybride2e}. The CTC model uses Markov assumptions to efficiently solve sequential problems by dynamic programming \cite{e2e-rnn-asr-2014,e2e-rnn-asr-2014-firstresult}, and the attention model uses an attention mechanism \cite{attention-mechanisum} to perform alignment between acoustic frames and labels. The hybrid CTC/attention model improves the robustness and achieves fast convergence, mitigates the alignment issues, and achieves comparable performance as compared to the conventional ASR based on a hidden Markov model (HMM)/deep neural networks (DNNs) \cite{hybride2e}. However, E2E model requires sufficiently large amount of paired speech-text data to achieve comparable performance \cite{comparison-e2e-s2s-asr-2017,e2e_eng_man}. The paired data is expensive, especially for low resource languages. There are huge amount of free unpaired speech-text data on the Internet, which we could make use of it with limited paired data to improve the E2E ASR in a semi-supervised manner.\\
\indent Cycle-consistent adversarial networks (CycleGAN) has demonstrated better model generalization using the cycle-consistent loss and the identity mapping loss on unpaired data \cite{CycleGAN2017}. Most of the studies, in the field of semi-supervised E2E ASR, exploit cycle-consistent loss to leverage unpaired data by combining speech-to-text and text-to-speech or text-to-text models \cite{cycle-consistence-e2e5,cycle-consistence-e2e2,cycle-consistence-e2e3,semie2e,cycle-consistence-e2e,asrtts}. However, there is no investigation into the effect of the identity mapping loss on semi-supervised E2E ASR performance while Zhu et al. observe that the identity mapping loss helps preserve the color of the input painting \cite{CycleGAN2017}. Besides, among the previously mentioned studies, an interesting work proposes the inter-domain loss, that targets at the extraction of an intermediate shared representation of speech and text using a shared network. This work combines speech-to-text and text-to-text mappings through the shared network in a semi-supervised end-to-end manner and thus improves the speech-to-text performance \cite{semie2e}. However, the inter-domain loss, which is the dissimilarity between the embedding from unpaired speech and text, might introduce errors to the shared network because it tries to minimize the distance between unpaired encoded speech and text. For instance, if the speech is ”\textit{actually the word I used there was presented}” and the text is ”\textit{what's wrong with that}”, the shared network learns to generate similar embedding for both of them.\\
\indent In this paper, we contribute to the previous work in the following aspects: 1) To the best of our knowledge, we are the first to investigate into the effect of the identity mapping loss on semi-supervised E2E ASR performance; 2) We propose a cycle-consistent inter-domain loss, which is dissimilarity between encoded speech and hypothesis, in order to help the shared network learn better representation; 3) We combine the identity mapping loss and the cycle-consistent inter-domain loss in a single framework for semi-supervised E2E ASR and achieve noticeable performance improvement; 4) We provide the analysis on the ASR output and the visualization of inter-domain embedding from speech and text, which explains the reason of performance gain by our proposed method.

\section{METHOD}
\subsection{Semi-supervised E2E ASR}
\begin{figure}[!htb]
    \centering
    \includegraphics[scale=0.23]{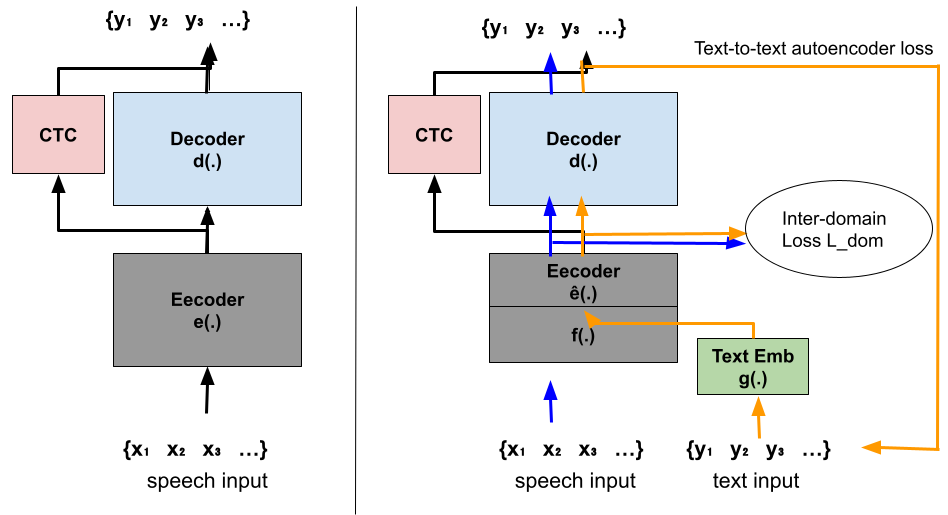}
    \caption{The architecture of hybrid CTC/attention model \cite{jointlyc2catt,hybride2e} (left) and the semi-supervised E2E model \cite{semie2e} (right).}
    \label{fig:architectures}
\end{figure}

Figure \ref{fig:architectures} (left) shows the architectures of hybrid CTC/atte-ntion model within the multi-task learning framework \cite{jointlyc2catt,hybride2e}, which is a encoder-decoder model. The encoder is trained by both CTC and attention objectives simultaneously. The encoder $e(.)$ transforms the acoustic feature $x=[x_1,x_2,...x_m]$ to the embedding $b=[b_1,b_2,...b_u]$, then the attention-based decoder $d(.)$ predicts the current label $y_{t}$ given the embedding $b$ and the previous label $y_{t-1}$. The processing pipeline is defined as follows \cite{hybride2e}:
%\vspace{-3mm}
\begin{align*}
%\footnotesize
    b=&e(x)\\
    [\Pr(y_{t}|y_{t-1},b),h_{t}]=&d(y_{t-1},h_{t-1},b)
\end{align*}
Where $y_{0}=\langle SOS \rangle$ is the start of a sequence label, and the initial state $h_0$ is zero. We write the above Eqs. as a sequence form \cite{hybride2e}
\begin{equation}
%\footnotesize
    d(b)=\Pr(y|b)=\prod_{t=1}^{|y|} \Pr(y_{t}|y_{t-1},b)
\end{equation}
Where $y = [y_1, y_2, . . . , y_{|y|}]$ is a predicted text, and $|y|$ is the length of text. The conventional loss for paired speech-text data $(x',y')\in Z$ is negative log likelihood of the ground-truth text $y'$ given the encoded speech $e(x')$ \cite{hybride2e}:
\begin{equation}
%\footnotesize
    L_{pair}=-\sum_{(x',y')\in Z} \log \Pr (y'|e(x'))
    \label{eq:loss_pair}
\end{equation}
For semi-supervised E2E, Karita et al. propose a framework which encodes speech and text into a common latent space ($\mathcal{B}$) and re-trains the E2E model on unpaired data by inter-domain loss \cite{unsupervised-i2i,unsupervised-nmt,unsupervised-mt-mono} and text-to-text autoencoder \cite{text-to-text-autoencoder} loss, see Figure \ref{fig:architectures} (right). We refer to the common intermediate embedding $b\in\mathcal{B}$  as an “inter-domain embedding”.  The input acoustic feature $x$ is fed to the encoder $e(.)=\hat{e}(f(.))$ and is transformed to the inter-domain embedding $b$. On the other hand, the input text is fed to the text embedding $g(.)$ and is processed by the shared encoder $\hat{e}(.)$, which generates the inter-domain embedding $b'=\hat{e}(g(y))$. 
The inter-domain loss is the dissimilarity between both embedding $b$ and $b'$, see the blue line in Figure \ref{fig:architectures} (right). Based on the author's code\footnote{https://github.com/ShigekiKarita/espnet-semi-supervised} and \cite{semie2e}, the author has explored adversarial loss \cite{gan2014},  Gaussian KL-divergence \cite{gaussiankl1959} and Maximum Mean Discrepancy (MMD) \cite{mmd} for the inter-domain loss, and we choose the one with the best result as the baseline for this work.\\
\indent The text-to-text autoencoder loss measures a negative log-likelihood that the encoder-decoder network can reconstruct text from unpaired text \cite{semie2e,text-to-text-autoencoder}, see the orange loop in Figure \ref{fig:architectures} (right), the definition is as follows: 
\begin{equation}
%\footnotesize
    L_{text}=-\sum \log \Pr (\textbf{y}|\hat{e}(g(\textbf{y})))
    \label{eq:loss_text_autoencoder}
\end{equation}

Since the inter-domain loss for speech-to-text plays a difficult role due to large difference between the speech and text domains, the objective consists of $L_{pair}$ and $L_{unpair}$ using tunable parameter $\alpha$ as follows \cite{semie2e}:
\begin{equation}
%\footnotesize
    L=\alpha L_{pair}+ (1-\alpha)L_{unpair}\label{eq:semie2e}
\end{equation}
Note that $L_{pair}$ is calculated on small paired data and $L_{unpair}$ is calculated on larger unpaired data. $L_{unpair}$ is composed of inter-domain loss and text-to-text autoencoder loss using tunable speech-to-text ratio $\beta$ as follows \cite{semie2e}:
\begin{equation}
    L_{unpair}=\beta L_{dom} + (1-\beta) L_{text}\label{eq:semie2e_unpair}
\end{equation}

\subsection{Semi-supervised E2E ASR using CycleGAN losses}
\begin{figure}[htb]
    \centering
    \includegraphics[scale=0.3]{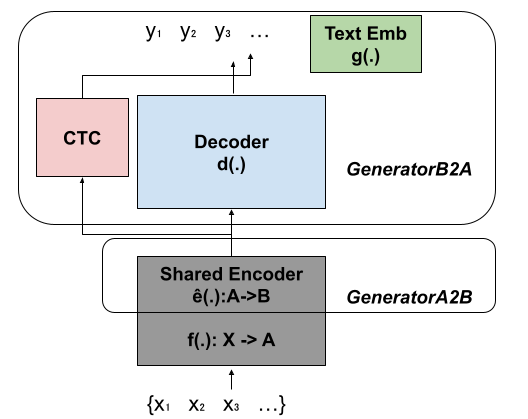}
    \caption{Illustration of viewing some components in semi-supervised E2E ASR as GeneratorA2B and GeneratorB2A. Note that the $g(.)$ is moved next to the Decoder is for the purpose of better understanding.}
    \label{fig:ill_generator}
\end{figure}
CycleGAN exploits cycle consistency loss and identity mapping loss to learn two mappings, $G:A\rightarrow B$ and $F:B\rightarrow A$, on unpaired data \cite{CycleGAN2017}. In the context of semi-supervised E2E, the shared encoder $\hat{e}(.)$ could be viewed as the generator $G:A\rightarrow B$ and the composition of decoder and text embedding $d(g(.))$ could be seen as the generator $F:B\rightarrow A$, see Figure \ref{fig:ill_generator}.\\

\begin{figure*}[htb!]
    \centering
    \subfloat[\centering The cycle-consistent inter-domain loss]{{\includegraphics[scale=0.4]{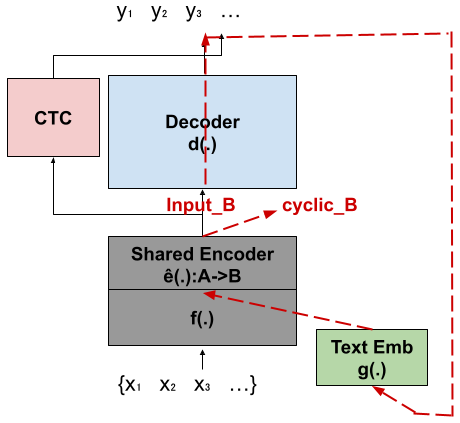} }}%
    \qquad
    \subfloat[\centering The identity mapping loss]{{\includegraphics[scale=0.4]{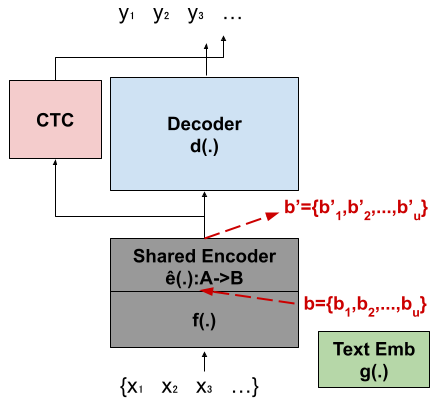} }}%
    \caption{Illustration of semi-supervised E2E ASR using CycleGAN losses. Note that the inter-domain embedding $b$ is either from speech or text. }%
    \label{fig:cycle_loss_and_idt_loss}%
\end{figure*}

\indent \underline{The identity mapping loss} encourages $G(b)=b$ and $F(a)=a$ to ensure that embedding retain their identity after translation \cite{CycleGAN2017}. Zhu et al. observe that the generators $G$ and $F$ are free to change the tint of input images without the identity mapping loss.
To make the shared encoder $\hat e(.)$ able to preserves important features after translation, the identity mapping loss is defined as follows (Figure \ref{fig:cycle_loss_and_idt_loss} (b)):
\begin{equation}
    L_{idt}=\rVert \hat{e}(b)-b \rVert_{1} \label{eq:idtloss}
\end{equation}
Where $b\in\mathcal{B}$ is the inter-domain embedding either from speech or text.\\
\indent \underline{Improving inter-domain loss using cycle-consistent loss} As mentioned in Section 1, the inter-domain loss might introduce errors to the shared encoder because it tries to minimize the distance between the inter-domain embedding from unpaired speech and text. To improve the inter-domain loss, we adopt the idea of cycle-consistent loss, which reconstructs input features. Figure \ref{fig:cycle_loss_and_idt_loss} (a) shows how we fuse the cycle-consistent loss to the inter-domain loss. The input\_B is the inter-domain embedding from speech and the cycle\_B is the reconstructed inter-domain embedding using the hypothesis of speech. The cycle-consistent inter-domain loss is defined as follows:
\begin{equation}\label{eq:cyc_dom}
\begin{split}
     L_{cyc,dom} &= \mathcal{D}(input\_B, cycle\_B)\\
     &= \mathcal{D}(e(x), \hat e(g(d(e(x)))))
\end{split}
\end{equation}
Where $\mathcal{D}(.)$ is a method to measure the distance between distributions. In this work, we use Maximum Mean Discrepancy (MMD), which achieves the best result with our propose method. The MMD is defined by a feature map $\phi: \rightarrow \mathcal{}{H} $, where $\phi$ is what's called a reproducing kernel Hilbert space \cite{mmd}. The definition is as follows:
\begin{equation}
    \mathcal{D}(P,Q)={\lVert \mathbb{E_{X \sim P}[\phi(X)]}-\mathbb{E_{Y \sim Q}[\phi(Y)]}\rVert}_{\mathcal{H}}
    \label{eq:mmd}
\end{equation}
Where $P$ and $Q$ is the distributions of inter-domain embedding from speech and text, respectively. $P$ and $Q$ are over $\mathcal{B}$.\\
\indent \underline{The proposed objective for the unpaired data} is the combination of the identity mapping loss, cycle-consistent inter-domain loss and text-to-text autoencoder loss using tunable speech-to-text ratio ($\beta$) , so the $L_{unpair}$ is adapted to:
\begin{equation} \label{eq:my_unpairloss_n}
\begin{split}
L_{unpair} & =\beta L_{unpair,speech} + (1-\beta) L_{unpair,text} \\
&=\beta (L_{cyc-dom}+L_{idt}(x))\\ &+(1-\beta)(L_{text}+L_{idt}(y))
\end{split}
\end{equation}
Where $\beta\in\{0,1\}$, and $x$ and $y$ are acoustic feature sequence and label sequence, respectively.

 \begin{table*}[htb!]
    \footnotesize
    \centering
    \begin{tabular}{|l|l|l|c|c|}
    \hline
    Model     & Objective & Remark & paired data & unpaired data\\
    \hline
    Initial model & Eq. (\ref{eq:loss_pair})&$L=L_{pair}$&v&\\
    \hline
    Semi-supervised models &Eq. (\ref{eq:semie2e})&\multicolumn{3}{c|}{$L=\alpha *L_{pair}+(1-\alpha)*L_{unpair}$}\\
    \hline
    -Baseline &Eq. (\ref{eq:semie2e}), (\ref{eq:semie2e_unpair}) & $L_{unpair}=\beta L_{dom}+(1-\beta)L_{text}$ &v&v\\
    -Retrain-idt &Eq. (\ref{eq:semie2e}), (\ref{eq:idtloss})& $L_{unpair}=L_{idt}$&v&v\\
    -Retrain-cyc &Eq. (\ref{eq:semie2e}), (\ref{eq:cyc_dom}), (\ref{eq:loss_text_autoencoder})& $L_{unpair}=\beta L_{cyc,dom}+(1-\beta)L_{text}$&v&v\\
    -Retrain-cyc+idt &Eq. (\ref{eq:semie2e}), (\ref{eq:my_unpairloss_n})&$L_{unpair}=\beta (L_{cyc,dom}+L_{idt}(x))+(1-\beta)(L_{text}+L_{idt}(y))$&v&v\\
    %-Retrain-cyc(2)+idt &Eq. (\ref{eq:semie2e}), (\ref{eq:my_unpairloss_2})&$L_{unpair}=\beta (L_{cyc,dom}+L_{idt})+(1-\beta)(L_{text}+L_{idt})$&v&v\\
    \hline
    \end{tabular}
    \caption{Experiment Terminology. Note that the small paired data for semi-supervised training is exactly the same data for initial model training.}
    \label{tab:experment_term}
\end{table*}
%\begin{comment}
%Besides, we want to enhance the text embedding network $g(.)$, so adapt the text-to-text loss to embedding-to-embedding loss as follows:
%\begin{equation}
 %   L_{text,emb}=\rVert g(y) - g(d(\hat{e}(g(y)))) \lVert_{1}\label{eq:emb2embloss}
%\end{equation}
%Therefore, another objective on unpaired data is as below:
%\begin{equation}
%\footnotesize
%    L_{unpair}=\beta (L_{cyc-dom}+L_{idt})+(1-\beta)(L_{text,emb}+L_{idt})
%    \label{eq:my_unpairloss_2}
%\end{equation}
%\end{comment}
%In the Section 4, we compare the performance of models, which are trained using different $L_{unpair}$ as shown in Tabel \ref{tab:experment_term}. 
For the simplicity, in the reminder of this paper, we refer to "the identity mapping loss" as $L_{idt}$ and "cycle-consistent inter-domain loss" as $L_{cyc,dom}$ and "text-to-text autoencoder loss" as $L_{text}$.

\section{EXPERIMENTAL SETUP}
\subsection{Resource}
We conduct experiments on three datasets: The first dataset is \underline{WSJ}, which contains 80 hours read speech with texts drawn from WSJ news text. It has a small 15-hour dataset (train\_si84) and the entire 80-hour dataset (train\_si284) as its official training datasets. The 64K vocabulary dev93 and eval92 are used for development and evaluation \cite{wsj}. The second dataset is \underline{LibriSpeech}, which has 1000 hours read speech with text derived from read audiobooks from the LibriVox project, and it has two splits of clean speech training data (test\_clean\_100 and test\_clean\_360) and other speech training data (test\_other\_500). In this paper, we only use test\_clean\_100 and test\_clean\_360 for semi-supervised training, and dev\_clean and test\_clean is for development and evaluation, respectively. The third dataset is \underline{Voxforge}, which consists of samples recorded and submitted by users using their own microphone \cite{Voxforge.org}. It has 8 languages (de,en,es,fr,it,nl,pt,ru) and each language has small train set ($<50$ hours), development and evaluation set. We use subset of train set for paired data and the rest is for unpaired data.

We use Espnet1 \cite{espnet} to build E2E ASR. The model is trained on WSJ using default setting, which is three layers Vgg \cite{vgg} bidirectional long short-term memory with projection (Vggblstmp) encoder with 1024 units and projections. The location based attention-decoder is one layer long short-term memory (LSTM) with 1024 units. The model for Librispeech is five layers vggblstmp encoder with 1024 units and projections and two layers location based attention-decoder, and the one for Voxforge is two layers vggblstmp encoder with 320 units and projections and one layer location based attention-decoder with 320 units. The text embedding $g(.)$ encodes the label, which is over $Y=\{'<SOS>', 'a', 'b',...\}$ , to an one-hot vector and process it by one layer BLSTM. Note that we do not use subword technique in this work. The shared encoder $\hat{e}(.)$ is the last layer of encoder $e(.)$.
The input acoustic feature is 80-bin log-mel filterbank with 3 pitch coefficients. The optimizer is adadelta and the batch size is 30. For decoding, we used a beam search algorithm with beam size of 20. The supervise loss ratio $\alpha$ in Eq. (\ref{eq:semie2e}) is 0.5 because it achieve the best result \cite{semie2e}.
Our codes\footnote{https://github.com/chiayuli/semi-supervised-E2E-using-CycleGAN.git} is accessible on github.
\subsection{Training pipeline}
The pipeline consists of four stages: First, the acoustic features and the label sequences are prepared; Second, the initial model is trained by Eq. (\ref{eq:loss_pair}) on small paired data and decoding on evaluation set; Third, the initial model is re-trained by Eq. (\ref{eq:semie2e}) on larger unpaired data. Note that we explore different $L_{unpair}$ shown in Table \ref{tab:experment_term} to investigate the effect on the ASR performance; Fourth, the recurrent neural network language model (RNNLM) is trained on unpaired text; Fifth, the retrain model is decoding without or with RNNLM using shallow fusion on evaluation set .

\section{RESULT AND DISCUSSION}
\subsection{Impact of using CycleGAN losses}
\begin{figure}[htb!]
    \includegraphics[scale=0.39]{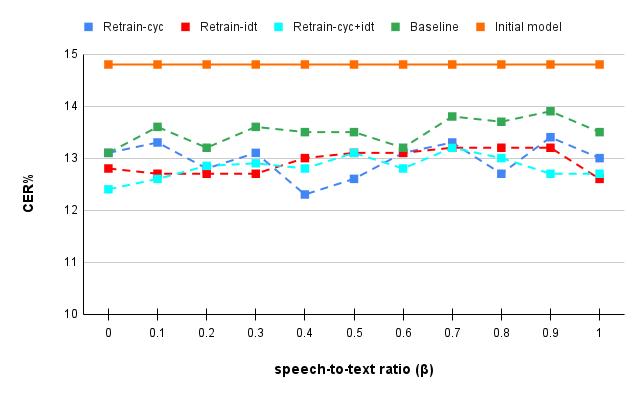}
    \caption{ASR performance (CER) on the WSJ eval92 versus the speech-to-text ratio ($\beta$).}%
    \label{fig:cer_comparison_wsj_eval92}%
\end{figure}

Figure \ref{fig:cer_comparison_wsj_eval92} shows CER on the WSJ eval92 while speech-to-text ($\beta$) varies. We compare Baseline, Retrain-idt, Retrain-cyc and Retrain-cyc+idt models. The CERs from initial model (14.8\%) and oracle (4.3\%), which is trained on the entire data in a supervised manner, is the upper bound and lower bound for retrained models, respectively.
The experimental result shows that Retrain-idt (red dots) has better CER than Baseline, and its performance does not fluctuate over speech-to-text ratio. That is to say, $L_{idt}$ helps model to achieve great performance when training on speech, text or both. The result also shows that the Retrain-cyc (blue dots) achieves the best CER at $\beta=0.4$ and it also performs better than the Baseline (green dots) all the time except at $\beta=0$. Besides, Retrain-cyc outperforms Baseline at $\beta=1$, which implies that the encoder using our proposed $L_{cyc,dom}$ generates better embedding than the one using $L_{dom}$.
Finally, the Retrain-cyc+idt (cyan dots), which combines $L_{idt}$, $L_{cyc,dom}$ and $L_{text}$, have advantages from the both losses and achieves good performance while $\beta$ varies.

\begin{comment}
\begin{figure*}[htb!]
    %\subfloat[\centering The Comparison of the models using $L_{idt}$,  $L_{cyc,dom}$, or both.]
    {{\includegraphics[scale=0.5]{Pictures/different_loss_result3.png}}}\hfill
    %\qquad
    %\subfloat[\centering The Comparison of the models using Eq.(\ref{eq:my_unpairloss_1}) or Eq. %(\ref{eq:my_unpairloss_2}) for $L_{unpair}$]
    %{{\includegraphics[width=0.5\linewidth]{Pictures/different_loss_result4.png}}}\hfill
    \caption{ASR performance (CER) on the WSJ eval92 versus the speech-to-text ratio ($\beta$).}%
    \label{fig:cer_comparison_wsj_eval92}%
\end{figure*}
\end{comment}

\subsection{Character/word error rate across corpus}
Table \ref{tab:cer_wer_wsj} and \ref{tab:cer_wer_librispeech} show the character/word error rate (CER/WER) on the WSJ eval92 and the LibriSpeech test\_clean without or with RNNLM, respectively. Note that the RNNLM is character based and only trained on the unpaired text. The "Type" in the table is related to the speech-to-text ratio ($\beta$). The settings $\beta=0$, $\beta=1$, $\beta \in [0.1,0.9]$ refers to "Text", "Speech" and "Both", respectively. \\
\indent Table \ref{tab:cer_wer_wsj} shows that Retrain-cyc+idt outperforms Baseline over all types. It achieves 8\% character error rate reduction (CERR) (6.8\% word error rate reduction (WERR)) at type "Both" as compared to Baseline and improves the initial models by 18.4\% CERR (15.4\% WERR). With RNNLM, Retrain-cyc+idt outperforms Baseline by 7.2\% CERR (3.9\% WERR) at type "Both" and improves the initial model by 20\% CERR (15.8 \% WERR). Table \ref{tab:cer_wer_librispeech} shows that Retrain-cyc+idt outperforms Baseline over all types. It improves Baseline at most by 4.9\% CERR (4.6 \% WERR) and the initial model by 7.4 \%CERR (6 \%WERR). With RNNLM, our approach outperforms Baseline by 3\% CERR (2.6\% WERR) and improves the initial models by 6\% CERR (5.9\% WERR).\\ 
\indent Table \ref{tab:voxforge_result} shows the CER on the evaluation sets for Italian(it), Dutch (nl), German (de), French (fr) in Voxforge. We constrain the paired data to be at least five hours and one third of the entire set because we want 1) the initial model has acceptable performance and 2) the split is aligned with the previous setting (WSJ and Librispeech uses $20\%\sim30\%$ of the entire data for paired data). The result shows that Retrain-cyc+idt outperforms baseline and it achieves noticeable character error rate reduction on four languages which contain very small paired data.

\begin{table}[htb!]
    \small
    \caption{ASR performance (CER/WER) on the WSJ eval92 w/o or w/ RNNLM.}
    \label{tab:cer_wer_wsj}
    \centering
    \begin{tabular}{lcccc}
    %\hline
    %\multicolumn{5}{c}{WSJ-SI84(paired) + WSJ-SI284(unpaired)}\\
    \hline
    \textbf{Model}&\textbf{Type}&\textbf{LM}&\textbf{CER(\%)} & \textbf{WER(\%)}\\
    \hline
    %initial model&&N&14.8&42.6\\\hline
    Oracle & - &N& 4.3&14.1\\
    Initial model& - &N& 14.8&42.6\\\hline
    Baseline &Text&N&13.1&38.3\\
    Retrain-cyc+idt &Text&N&\textbf{12.4}&\textbf{36.9}\\\hline
    Baseline &Speech&N&13.5&39.0\\
    Retrain-cyc+idt &Speech&N&\textbf{12.7}&\textbf{37.5}\\\hline
    Baseline &Both&N&13.5&39.6\\
    Retrain-cyc+idt &Both&N&\textbf{12.5}&\textbf{36.9}\\\hline\hline
    Oracle & - &Y& 2.3&4.9\\
    Initial model& - &Y& 8.3&17.6\\\hline
    Baseline &Text&Y&7.3&15.8\\
    Retrain-cyc+idt &Text&Y&\textbf{7.1}&\textbf{15.4}\\\hline
    Baseline &Speech&Y&7.3&16.7\\
    Retrain-cyc+idt &Speech&Y&\textbf{7.0}&\textbf{15.1}\\\hline
    Baseline &Both&Y&7.4&15.8\\
    Retrain-cyc+idt &Both&Y&\textbf{6.9}&\textbf{15.2}\\
    \hline
    \end{tabular}
\end{table}
\begin{table}
    \small
    \caption{ASR performance (CER/WER) on the LibriSpeech test\_clean w/o or w/ RNNLM. (These models are trained on 100+360 hours train\_clean set.)}
    \label{tab:cer_wer_librispeech}
    \centering
    \begin{tabular}{lcccc}
    \hline
    \textbf{Model}&\textbf{Type}&\textbf{LM}&\textbf{CER(\%)}&\textbf{WER(\%)}\\
    \hline
    Oracle &-&N&3.9&11.0\\
    Initial model&-&N&8.7&22.7\\
    \hline
    Baseline &Text&N&8.5&22.4\\
    Retrain-cyc+idt &Text&N&\textbf{8.3}&\textbf{21.7}\\
    \hline
    Baseline &Speech&N&8.5&22.3\\
    Retrain-cyc+idt &Speech&N&\textbf{8.1}&\textbf{21.2}\\
    \hline
    Baseline &Both&N&8.5&22.4\\
    Retrain-cyc+idt &Both&N&\textbf{8.1}&\textbf{21.4}\\\hline\hline
    Oracle &-&Y&3.5&8.9\\
    Initial model&-&Y&7.0&16.1\\
    \hline
    Baseline &Text&Y&6.8&15.8\\
    Retrain-cyc+idt &Text&Y&\textbf{6.7}&\textbf{15.6}\\
    \hline
    Baseline &Speech&Y&6.7&15.6\\
    Retrain-cyc+idt &Speech&Y&6.7&\textbf{15.5}\\
    \hline
    Baseline &Both&Y&6.8&15.6\\
    Retrain-cyc+idt &Both&Y&\textbf{6.6}&\textbf{15.2}\\
    \hline
    \end{tabular}
\end{table}
\iffalse
\begin{table}[htb!]
  \footnotesize
  \centering
  \caption{Character error rate reduction (CERR) on Voxforge evaluation. Note that we exclude the languages (i.e. pt, es, ru), whose "All supervised model" performs badly in E2E ASR. }
    \begin{tabular}{c|ccccc}
        \hline
        \textbf{lang.} &  \textbf{paired/All}& \textbf{All supervised.} &\textbf{Baseline}& \textbf{Retrain-cyc+idt}\\
              &\textbf{(hour)}& \textbf{CER(\%)}&   \textbf{CERR(\%)} & \textbf{CERR(\%)}\\\hline
         it&5/15&12.7&29.8&43.5\\
         nl&5/8&20.3&25.2&32.4\\
         de&11/45&20.3&4.1&5.2\\
         fr&5/19&30.8&16.8&28.7\\
         %es&&&&&\\
         %ru&&12&&&\\
         \hline
    \end{tabular}
    \label{tab:voxforge_result}
\end{table}
\fi
\begin{table}[htb!]
\small
  \centering
  \caption{ASR performance (CER) on Voxforge evaluation. Note that we exclude the languages (i.e. pt, es, ru) because their oracle and baseline perform badly. }
    \begin{tabular}{l|ccccc}
        \hline
        \textbf{Models} &  \textbf{it}& \textbf{nl} &\textbf{de}& \textbf{fr}\\
        \textbf{paired data (hour)}&5&5&10&5\\\hline
         Oracle&12.9&25.2&5.6&30.8\\
         Initial model&29.4&35&20.3&53.3\\
         Baseline&22.1&33.7&20.2&47.9\\
         Retrain-cyc+idt&\textbf{19.7}&\textbf{32.8}&\textbf{19.4}&\textbf{41.4}\\
         %es&&&&&\\
         %ru&&12&&&\\
         \hline
    \end{tabular}
    \label{tab:voxforge_result}
\end{table}

\raggedbottom
\subsection{The number of layers in the shared encoder $\hat{e}(.)$}
\label{subsection:num_of_layers_in_shared_encoder}
Table \ref{tab:shared_encoder} shows the CER on the WSJ eval92 when increasing the number of layers in the shared encoder $\hat{e}(.)$. When the Type is "Text", increasing the number of layers has CER degradation, however, the CERs get improved at type "Both" and "Speech". The CER degradation at "Text" might be due to over-fitting, which implies that text contains less information than speech (speaker characteristic, prosody), so it does not require complicated network to model it.  On the contrary, the inter-domain embedding requires relatively complicated network to model it (because both speech and text domains are quite different), so the CER show noticeable reduction while increasing the number of layers (network parameters) in the shared encoder.
\begin{table}[htb]
    \small
    \caption{The CER(\%) on the WSJ eval92 using different number of layers in the shared encoder $\hat{e}(.)$ .}
    \label{tab:shared_encoder}
    \centering
    \begin{tabular}{cccc}
    \hline
    \textbf{Num. of layers}&\textbf{Text}&\textbf{Both(Speech+Text)}&\textbf{Speech}\\
    %\textbf{in $\hat{e(.)}$}&\textbf{CER(\%)}&\textbf{CER(\%)}&\textbf{CER(\%)}\\
    \hline
    default(1)&12.6&12.8&12.7\\
    3&12.8&12.5&12.6\\
    4&13.2&\textbf{12.2}&12.6\\
    \hline
    \end{tabular}
\end{table}
\raggedbottom
\section{ANALYSIS}
\subsection{ASR output, substitution, deletion and insertion}
Table \ref{tab:sub_del_ins_wer} shows the substitution, deletion, and insertion at word level on WSJ eval92 without RNNLM. The result shows that Retrain-idt reduces insertions significantly. $L_{idt}$ mitigates the problem of recognizing /sil/ as character or predicting word boundary wrongly in Baseline, see Table \ref{tab:example_1}.
Retrain-cyc+idt improves substitution significantly and insertions. We observe that Retrain-cyc+idt has better contextual relations and acoustic frame-to-character mapping. For instance, it predicts "trend" correctly, while the baseline predicts "tr\textcolor{red}{a}nd", 
and "received" versus "re\textcolor{red}{s}eived", "strength" versus "stre\textcolor{red}{in}th", and "department" versus "\textcolor{red}{at}partment".
\begin{table}[htb!]
%\footnotesize
    \centering
    \caption{The substitution, deletion and insertion at word level on WSJ eval92 w/o RNNLM. }
    \begin{tabular}{l|c|c|c}\hline
        Models & SUB&DEL&INS\\\hline
         Baseline &31.9&2.8&5.2 \\
         Retrain-idt&30.6&2.9&\textbf{4.7}\\
         Retrain-cyc&30.7&2.8&5.1\\
         Retrain-cyc+idt&\textbf{29.3}&2.8&5.0\\\hline
    \end{tabular}
    \label{tab:sub_del_ins_wer}
\end{table}
\begin{table}[htb!]
    \centering
    \caption{Words in REF and hypothesis from Baseline and retrained models}
    \begin{tabular}{l|lll}\hline
    REF & Baseline & Retrain-idt &Retrain-cyc\\\hline
    departed & \textcolor{red}{th}e parted& departed   & the parted\\ 
    commodore &commod \textcolor{red}{or} &commod\textcolor{red}{a}re & commod\textcolor{red}{a}re\\
    /sil/ & a&/sil/&/sil/\\
    making & mak\textcolor{red}{e at} &making& makean\\
    \hline
    \end{tabular}
    \label{tab:example_1}
\end{table}

\subsection{t-SNE visualization of inter-domain embedding}
\begin{figure}[htb!]
    \subfloat[\centering Baseline]
    {{\includegraphics[width=0.5\linewidth]{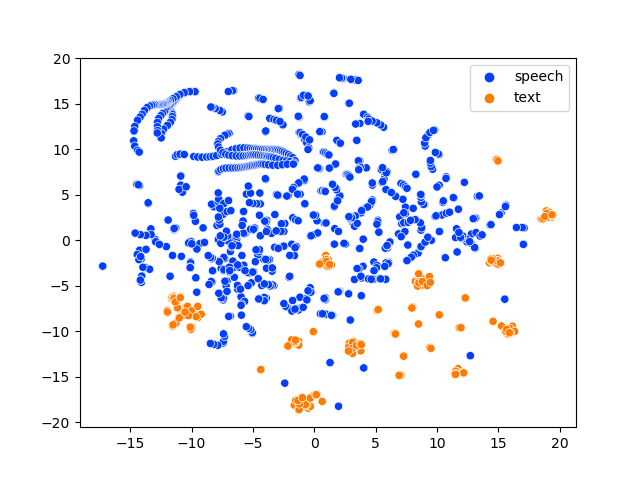}}}\hfill
    %\qquad
    \subfloat[\centering Retrain-idt]
    {{\includegraphics[width=0.5\linewidth]{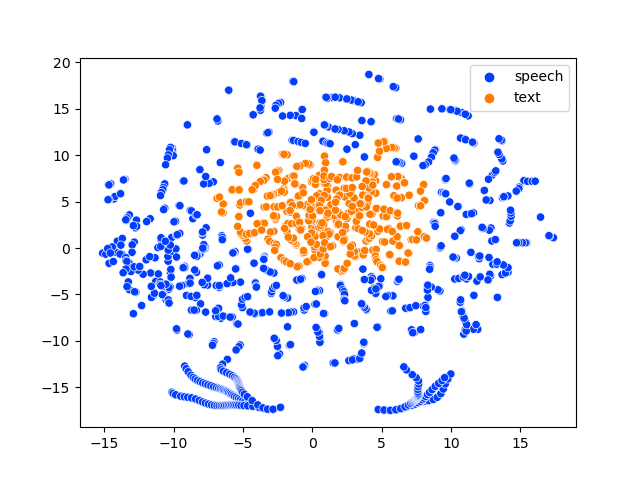}}}\hfill
    \subfloat[\centering Retrain-cyc]
    {{\includegraphics[width=0.5\linewidth]{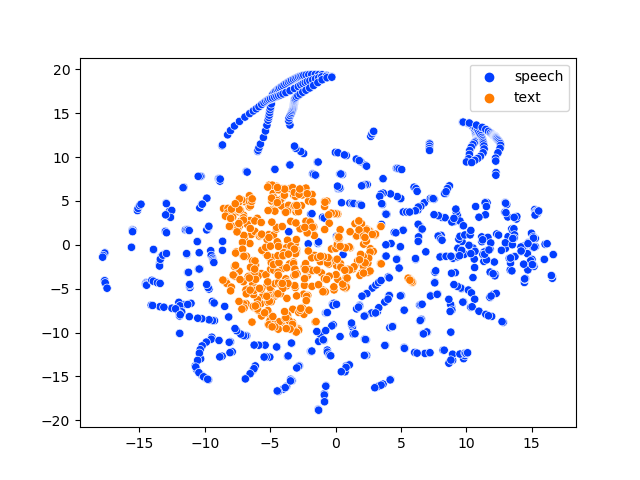}}}\hfill
    \subfloat[\centering Retrain-cyc+idt]
    {{\includegraphics[width=0.5\linewidth]{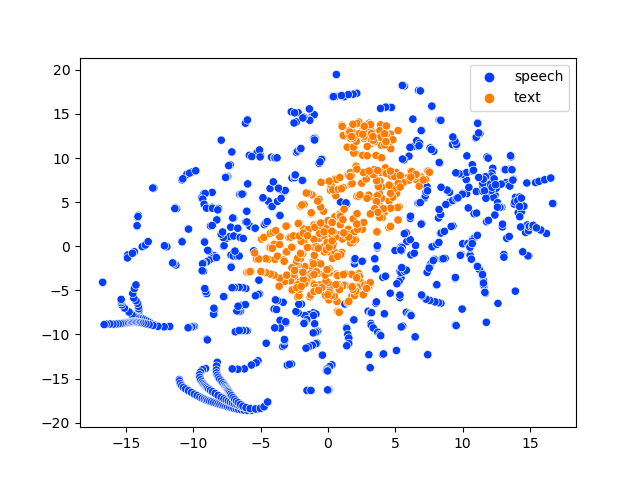}}}%
    \caption{t-SNE visualization of inter-domain embedding.}%
    \label{fig:tsne_correctTxt}%
\end{figure}
\begin{figure}
    \subfloat[\centering Baseline]
    {{\includegraphics[width=0.5\linewidth]{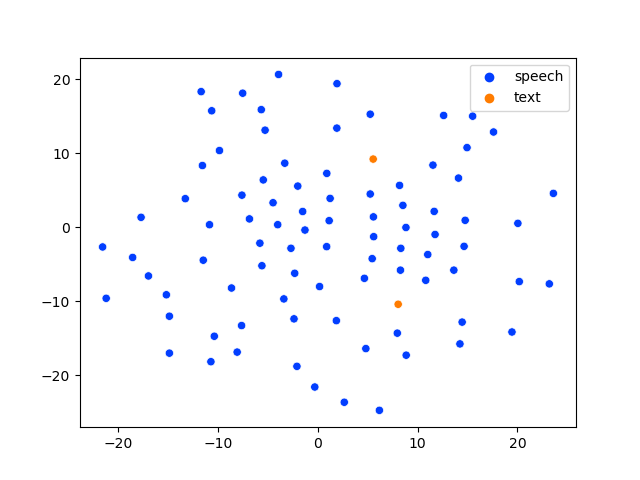}}}\hfill
    %\qquad
    %\subfloat[\centering Retrain-idt ]
    %{{\includegraphics[width=0.25\linewidth]{Pictures/v16.png}}}\hfill
    %\subfloat[\centering Retrain-cyc(1) ]
    % {{\includegraphics[width=0.25\linewidth]{Pictures/v15.png}}}\hfill
    \subfloat[\centering Retrain-cyc+idt ]
    {{\includegraphics[width=0.5\linewidth]{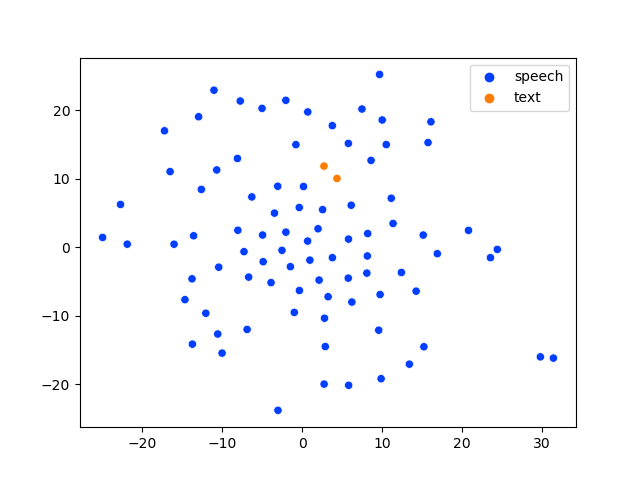}}}%
    \caption{t-SNE visualization of inter-domain embedding from the unpaired speech and text. Note that text contains OOV labels.}%
    \label{fig:tsne_wrongTxt}%
\end{figure}
We apply dimensional reduction to the two-dimensional plane using t-distributed stochastic neighbor embedding (t-SNE) \cite{tsne}.
Figure \ref{fig:tsne_correctTxt} shows that the inter-domain embedding from speech and text. Our retrained models show better regularization for speech and text than Baseline because the embedding is more mixed together. Besides, the embedding from text is the subset of the one from speech, which matches what we observed in section  \ref{subsection:num_of_layers_in_shared_encoder}: text features contains less information than speech. Figure \ref{fig:tsne_wrongTxt} shows the visualization of inter-domain embedding from one unpaired speech-text and text simply contains repeated Out-of-Vocabulary (OOV) labels (length=2). The result shows that the embedding of OOV labels from both models are not deviated too much from the speech. However, the embedding for two repeated OOV labels from Baseline (orange dots) are far from each other. Note that each orange dot represents one label (we get four orange dots after using four repeated labels). Our retrained models produces better embedding for OOV label in this case.

\section{CONCLUSION AND FUTURE WORK}
\label{sec:print}
In this study, we propose a novel method, which combines CycleGAN losses and inter-domain losses for semi-supervised E2E ASR, and show its effectiveness on the WSJ, LibriSpeech and Voxforge datasets. Our proposed method effectively utilizes advantages from both CycleGAN and inter-domain loss. Overall, our method improves the Baseline by 8\% CERR (6.8\% WERR) on the WSJ eval92, 4.9\% CERR (4.6\% WERR) on LibriSpeech test set, and  8.5\% CERR on Voxforge (avg. it, nl, de, fr) in a semi-supervised setting. In future work, we plan to improve our method by automatic speech-to-text ratio ($\beta$) tuning and extend it for fully unsupervised E2E ASR.
%clean condition by semi-supervised training on a 100-hour paired and 360-hour unpaired  LibriSpeech dataset.

\newpage

\bibliographystyle{IEEEbib}
\bibliography{strings,refs}

\end{document}